\documentclass{article}
\usepackage{spconf,amsmath,graphicx}

\title{WEIGHT-IMPORTANCE SPARSE TRAINING IN KEYWORD SPOTTING}
\name{Sihao Xue, Zhenyi Ying, Fan Mo, Min Wang, Jue Sun}
\address{NIO Co., Ltd\\
	      \emph{\{sihao.xue, zhenyi.ying, fan.mo, min.wang, jue.sun\}@nio.com }}
\begin{document}
\topmargin=0mm
%\ninept
%
\maketitle
\begin{abstract}
Large size models are implemented in recently ASR system to deal with complex speech recognition problems. The number of parameters in these models makes them hard to deploy, especially on some resource-short devices such as car tablet. Besides this, at most of time, ASR system is used to deal with real-time problem such as keyword spotting (KWS). It is contradictory to the fact that large model requires long computation time.

To deal with this problem, we apply some sparse algorithms to reduces number of parameters in some widely used models, Deep Neural Network (DNN) KWS, which requires real short computation time. We can prune more than 90 \% even 95\% of parameters in the model with tiny effect decline. And the sparse model performs better than baseline models which has same order number of parameters. Besides this, sparse algorithm can lead us to find rational model size automatically for certain problem without concerning choosing an original model size. 
\end{abstract}
\section{Introduction}
\label{sec:intro}
In recent years, great progress has been achieved in speech recognition. The main reason is the usage of a large neural network trained on large-scale datasets. People usually design networks with a large number of parameters to build a recognition model. However, this requires massive computation and memory capacity, which are limited to some machines which do not have strong computation ability and enough storage space such as car tablet. 

Most of the time, people use ASR system to deal with the real-time problem, long computation time is unacceptable. It is contradictory to the fact that sizeable neural network usually requires more time to run. Especially in footprint keyword spotting (KWS), small storage space, short computation time and low CPU usage are required.

Besides these, how to design an appropriate neural network architecture is a classic question in deep learning.  Most of the time, selecting some empirical architecture and adjusting parameters is usually implemented. This method often leads to big network than the true optimum one, resources waste and overfitting problem.

Motivated by this, we implemented sparse model on our KWS model. We apply several algorithms to prune the model. The final model has similar performance with original one but has thinner structure and requires less computational complexity.

%Services, Inc.: Phone +1-979-846-6800 or email
%to \\\texttt{dsw2018@cmsworkshops.com}.

\section{RELATED WORK}
\label{sec:format}

There have been plenty of research to reduce the network size by pruning the model. Li et al. [1] use sparse shrink model to prune a CNN model. Han et al. [2][3] prune the small-weight connections to prune a CNN model. Narang et al. [4][5] apply sparsity algorithm to prune RNN. Hassibi et al. [6]  and Yann et al. [7] uses Hessian-based approach to prune weight.

There are also plenty of literature on the topic of KWS. Offline Large Vocabulary Continuous Speech Recognition (LVCSR) systems can be used for detecting the keywords of interest. [8][9]. Moreover, Hidden Markov Models (HMM) are commonly used for online KWS systems [10][11]. In traditional, Gaussian Mixture Models (GMM) is used in acoustic modeling under the HMM framework. It is replaced by Deep Neural Network (DNN) with time goes on [12]. And several architectures have been applied [13][14]
\section{KEYWORD SPOTTING}
\label{sec:pagestyle}

KWS is the entrance of ASR. It provides interactive intention for subsequent recognition problem. In general, KWS works on local devices and processes voice data collected by the microphone so that short delay time and low memory storage are required to ensure user experience and acceptable consumption. The early KWS is based on offline continuous speech recognition with GMM-HMM [10][11]. With the great success of Deep Neural Network in continuous speech recognition, traditional GMM-HMM is replaced by DNN[12]. Recently, Chen et al. [15] design a KWS strategy without HMM.

In our research, we use finite state transducer (FST) to realize KWS by employing word unit. FST consists of a finite number of states. Each state is connected by transition labeled with input/output pair. Its states transition is depending on input and transition rules. For example, ``happy". First search "happy" in the dictionary for its phone units, which is ``HH AE1 P IY0". Then find its tri-phone such as h-ay1-p which may occur in voice data and do clustering to generate each state. During clustering, we let the tri-phones whose central phone is ``HH" and ``AE1"  as the first word, ``P" and ``IY0" as the second word. The "happy" FST is shown in Figure 1. The expression is input/output pair, the arrow means state transformation. Device wakes up when the output equals 1. It wakes up while ``happy" occurs. 

\begin{figure}[htb]

\begin{minipage}[b]{1.0\linewidth}
 \centering
  \centerline{\includegraphics[width=5cm]{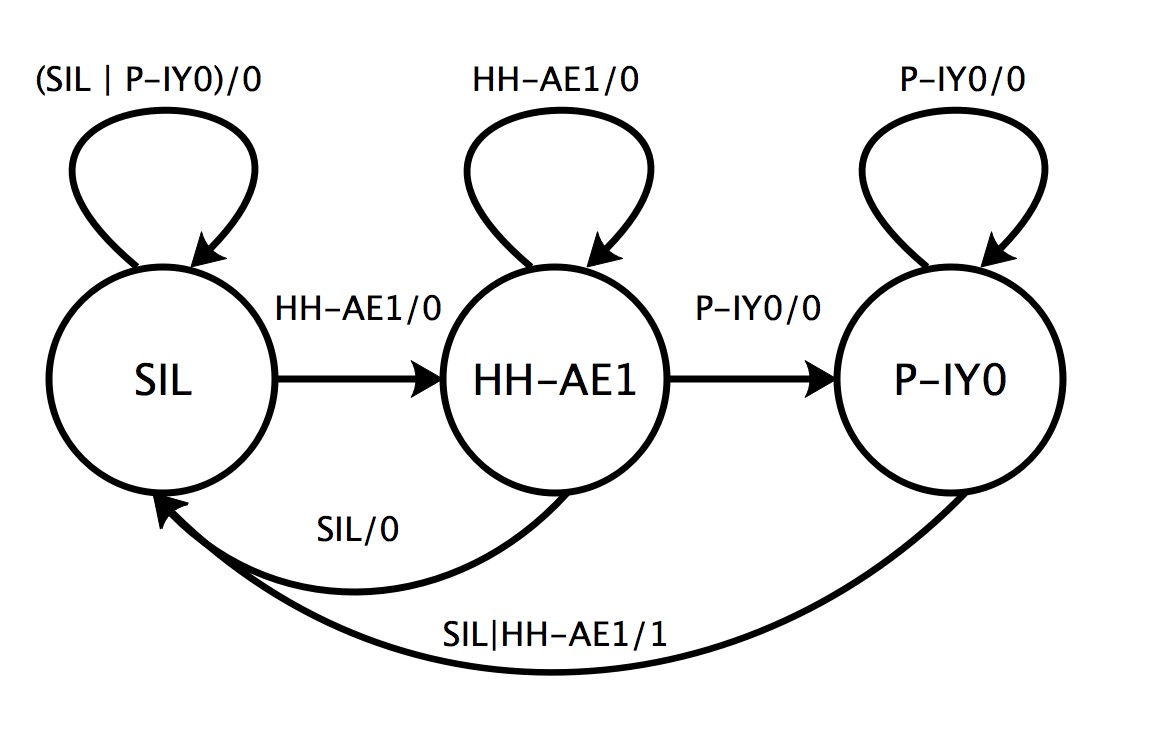}}
 % \vspace{2.0cm}
%  \centerline{(a) Result 1}\medskip
\end{minipage}

\caption{FST for keyword ``happy"}%
\label{fig:res}

\end{figure}

\section{SPARSE MODEL ALGORITHM}
\label{sec:majhead}

In this section, we elaborate how we implement the sparse algorithms in KWS problems.

\subsection{Pruning Algorithm}
\label{ssec:subhead}
 
\subsubsection{Pruning Based on Weight Magnitude}
\label{sssec:subsubhead}

The most simple and naive method is pruning the network based on weight magnitude. This algorithm assumes that small magnitude is corresponding to little importance, so we can delete some small weight with small magnitude. The topology is shown in Figure.2. For this algorithm, we have several different options to implements:\\

1. Delete certain proportion number of remain weights after several iterations.

2. Delete a certain number of weights after several iterations.

3. Delete weights whose magnitude is less than a certain threshold.

4. Delete weights whose magnitude is less than a certain threshold, but the percentage number of deleted weights must be less than a certain percentage.

\begin{figure}[htb]

%\begin{minipage}[b]{1.0\linewidth}
% \centering
%  \centerline{\includegraphics[width=8.5cm]{test.jpeg}}
 % \vspace{2.0cm}
 % \centerline{(a) Result 1}\medskip
%\end{minipage}

\begin{minipage}[b]{.48\linewidth}
  \centering
  \centerline{\includegraphics[width=3.5cm]{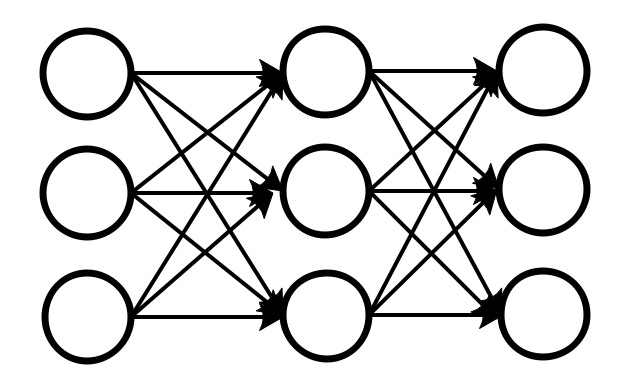}}
 % \vspace{1.5cm}
 \centerline{(1)  }\medskip
\end{minipage}
%!TEX encoding = UTF-8 Unicode%\hfill
\begin{minipage}[b]{0.48\linewidth}
 \centering
  \centerline{\includegraphics[width=3.5cm]{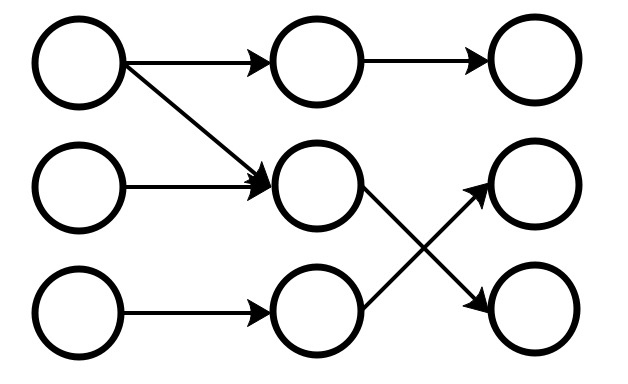}}
 % \vspace{1.5cm}
  \centerline{(2)  }\medskip
\end{minipage}

\caption{Sparse model topology structure bases on pruning based on weight magnitude. Network (1) is original fully connect network, network(2) is sparse model topology structure}%
\label{fig:res}

\end{figure}

%The structure is showed in Figure.2. Network (1) is original fully connect network, network (2) is spare model topology architecture.

Continuing the train after each pruning operation to let the network re-converge. Besides pruning algorithm, selection of learning rate after pruning influences final model performance. It is easy to imagine that learning rate has to be set relatively large because the model has changed significantly after one pruning iteration.

Actually, this algorithm is partly similar to normalization. This algorithm will reduce normalization value as its elimination on some weight parameters. 

However the assumption of this algorithm may lead to some problems. It is short of convincing to believe that less magnitude means less importance. It may be that some weight magnitude is small for its large input from the last layer. So this simple algorithm may destroy the neural network.

\subsubsection{Pruning Based on Affine Transformation Value}
\label{sssec:subsubhead}

For each node, its input is affine transformation output of last stage layer nodes:

\begin{equation}
	in_{i+1}  = W_{i, i+1}out_{i}
	\label{eq1}
\end{equation}

\begin{figure}[htb]

\begin{minipage}[b]{1.0\linewidth}
 \centering
  \centerline{\includegraphics[width=5cm]{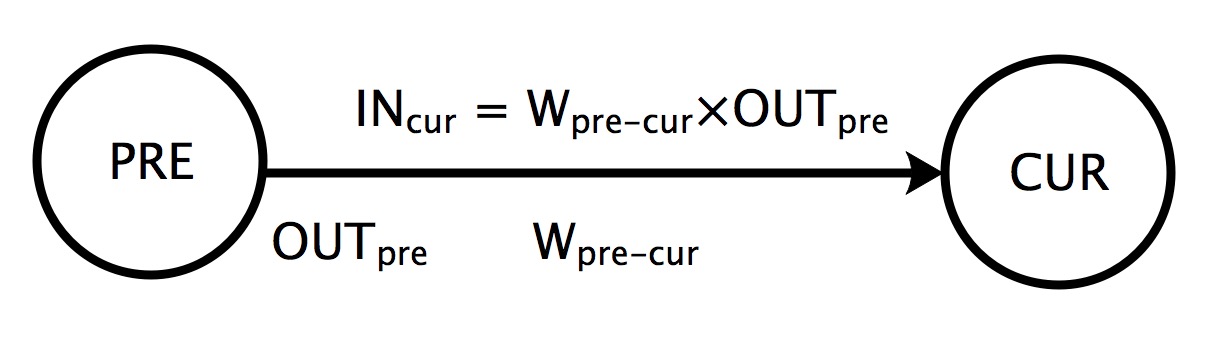}}
 % \vspace{2.0cm}
%  \centerline{(a) Result 1}\medskip
\end{minipage}

\caption{Compute process of each affine transformation weight.}%
\label{fig:res}

\end{figure}

Comparing with large magnitude input, neglecting small magnitude input is an obvious way of pruning the network. The computing process is shown in Figure.3. This algorithm has same options to implements as Pruning Based on Weight Magnitude method which is mentioned before. The difference is this method based on the affine transformation value. Also, it requires continuing the train after each pruning operation to let the network re-converge.

\subsubsection{Pruning Based on Dictionary Algorithm}
\label{sssec:subsubhead}
Dictionary algorithm aims to keep important weights in network only. it assumes that weight importance is its contribution for the whole network statistically. If the input dataset is D$_{input}$, it has N frames. The importance I$_{ij}$ of w$_{ij}$ is:

\begin{equation}
	I_{ij} = \sum_{k}^N|w_{ij}D_{input, k, i}|
	\label{eq1}
\end{equation}
where k is the index of input data.

After pruning less important weight. The dictionary algorithm revises the network by using important weights to evaluate the value of unimportant weights. For easy to compute, it revises the network only based on the unimportant one which has the greatest importance among all unimportant weights of one node. For example, for one node node$_i$, the importance of this node is I$_{ji}$, j is the index of last layer node. These value is divided into two set: important set \{I$_{j_{im}, i}$, I$_{j_{im}, i}$...\} and unimportant set \{I$_{j_{unim}, i}$, I$_{j_{unim}, i}$....\}, where im means important and unim means unimportant. 
Then find the unimportant weight with maximum importance:
\begin{equation}
	I_{j_{re}, i} = max_{k}\{I_{k_{unim}, i}\}
	\label{eq1}
\end{equation}
Then revising other weights according their importance:
\begin{equation}
	w_{ji} = w_{ji} + sign(w_{ji})w_{ji} \frac{I_{j_{re}, i}}{N_{im}I_{j_{im},i}}
	\label{eq1}
\end{equation}
where we assign the revision value to every important weight equally. The compute process is showed in Figure.4.

\begin{figure}[htb]

\begin{minipage}[b]{1.0\linewidth}
 \centering
  \centerline{\includegraphics[width=3cm]{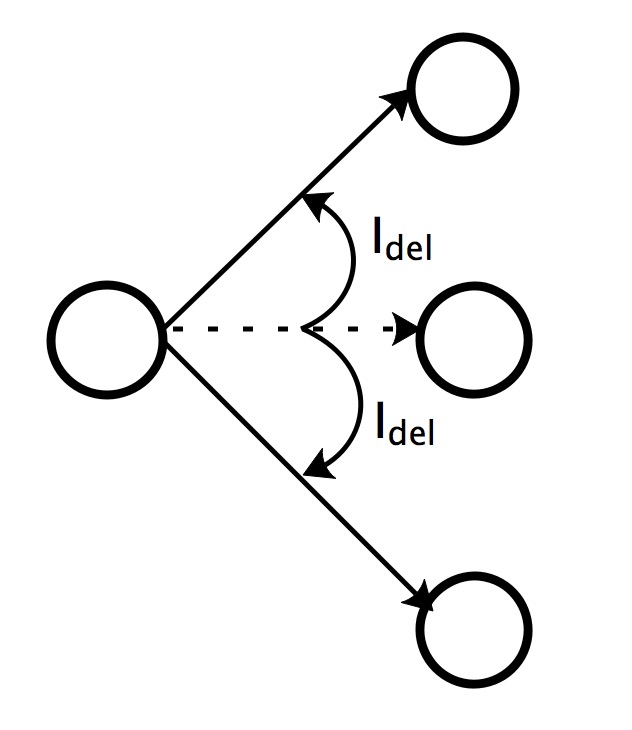}}
 % \vspace{2.0cm}
%  \centerline{(a) Result 1}\medskip
\end{minipage}

\caption{Compute process of 4.1.3. Full arrow is remaining weight, dot arrow is pruned weight. Pruned weight provide importance information to remain weight to revise.}%
\label{fig:res}

\end{figure}

\subsubsection{Pruning Based on Optimal Brain Damage}
\label{sssec:subsubhead}
This algorithm is proposed by Yann et al. [7]. Its main idea is based on the second-order derivation of loss. The assumption is introduced that $\Delta$E is caused by deleting each parameter individually. This assumption decreases the computation notable because it requires a large resource to compute Hessian Matrix and its inverse matrix.

\subsubsection{Pruning Based on Optimal Brain Surgeon}
\label{sssec:subsubhead}
This algorithm is proposed by Hassibi et al. [6]. The main difference between OBS and OBD is that OBS uses whole Hessian Matrix to prune the network. It requires a larger resource to compute and may not be suitable for the big models nowadays.\\

\noindent It is noteworthy that 4.1.2, 4.1.3, 4.1.4, 4.1.5 relies on training data. 4.1.2 and 4.1.3 require affine transformation value. 4.1.4 and 4.1.5 require loss value, so it is important to pay more attention to the distribution of input data. If the training data is bias, the pruning decision will be bias.

\subsection{Matrix Decomposition}
\label{ssec:subhead}
After pruning the network, it is a problem how to implement the sparse network in practice. The most straightforward idea is setting the pruned weight to zero. But neither store storage nor computation requirement is reduced by this method. Another thought is designing a map to connect related node. However, it may be difficult to realize and errors occurring probability is relatively high. We implement a compromising method, matrix decomposition, to deal with this awkward situation.

if we have two matrices whose size is $m \times r$ and $r \times n$ (the number of parameter is $(m + n) \times r$ ), the product of them is $m \times n$ (the number of parameters is $m \times n$). When we use $m \times r$ and $r \times n$ matrices to replace a $m \times n$ matrix, the store and computation requirements is reduced if the r is relative small.

To decompose a matrix, we design two networks showed in Figure.5.
\begin{figure}[htb]

%\begin{minipage}[b]{1.0\linewidth}
% \centering
%  \centerline{\includegraphics[width=8.5cm]{test.jpeg}}
 % \vspace{2.0cm}
 % \centerline{(a) Result 1}\medskip
%\end{minipage}

\begin{minipage}[b]{.48\linewidth}
  \centering
  \centerline{\includegraphics[width=2.0cm]{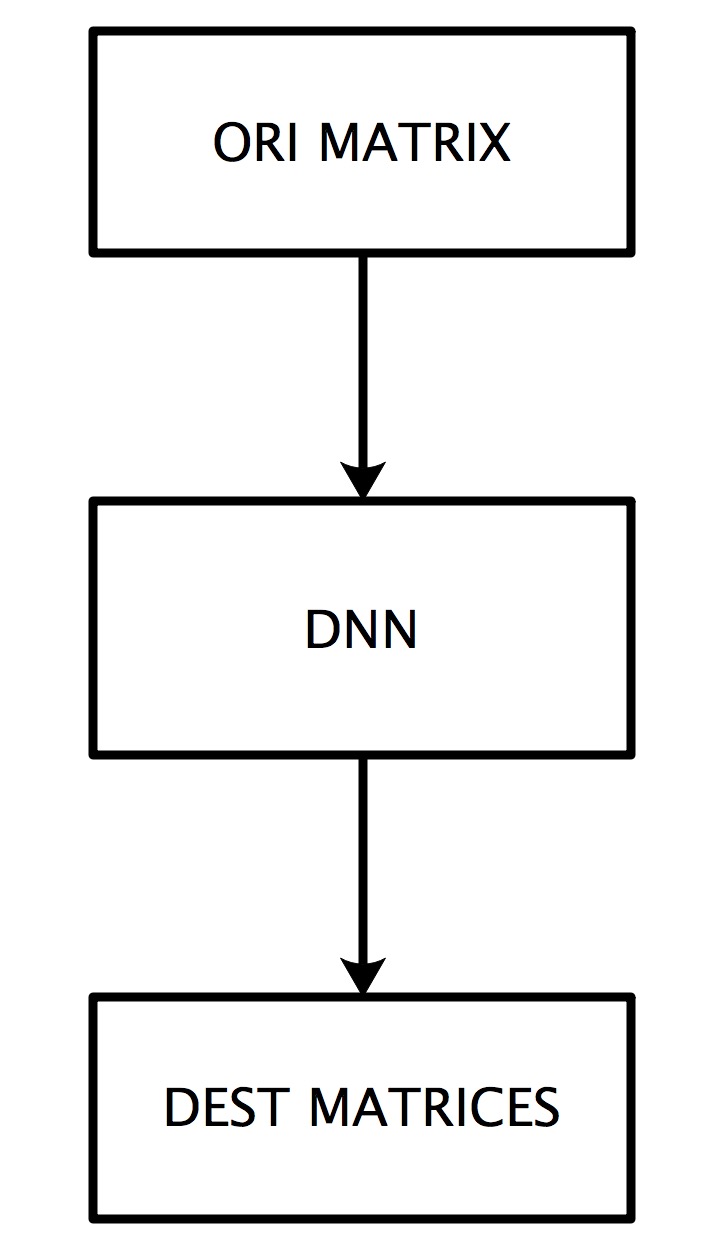}}
 % \vspace{1.5cm}
 \centerline{(1)  }\medskip
\end{minipage}
%!TEX encoding = UTF-8 Unicode%\hfill
\begin{minipage}[b]{0.48\linewidth}
 \centering
  \centerline{\includegraphics[width=2.0cm]{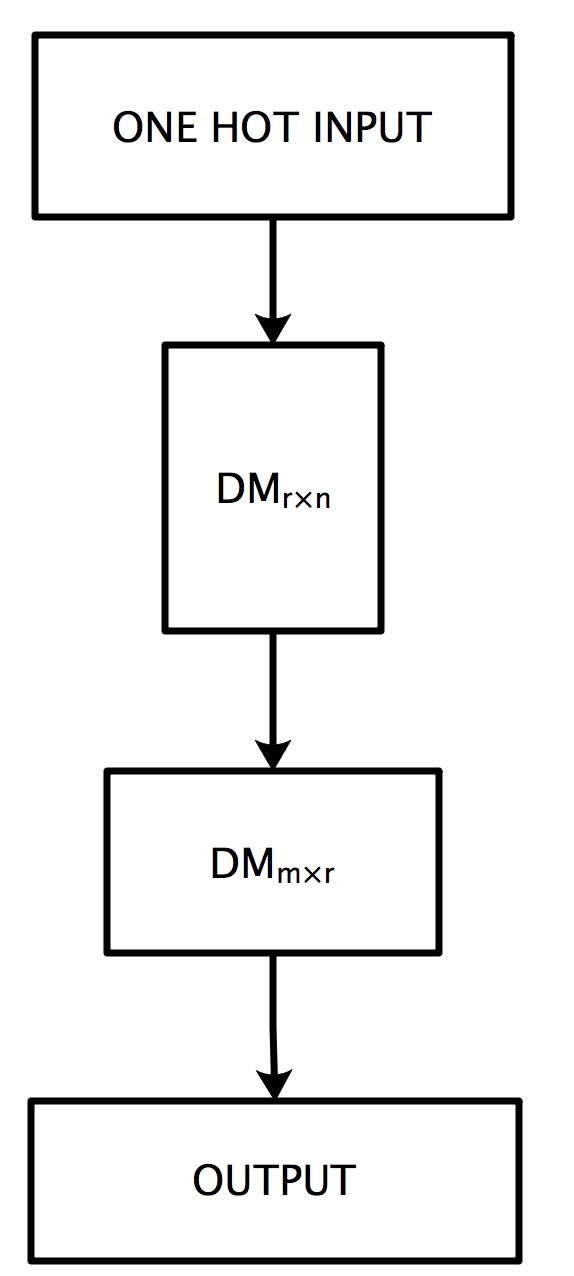}}
 % \vspace{1.5cm}
  \centerline{(2)  }\medskip
\end{minipage}

\caption{Two network architecture}%
\label{fig:res}

\end{figure}

Network(1) uses original matrix $M_{m \times n}$ as input and decomposed matrices $DM_{m \times r}$ and $DM_{r \times n}$ as output. The loss function can be defined as:

\begin{equation}
	%w_{ji} = w_{ji} + sign(w_{ji})w_{ji} \frac{I_{j_{re}, i}}{N_{im}I_{j_{im},i}}
	L  = |DM_{m \times r}DM_{r \times n}|^2
	\label{eq1}
\end{equation}

Network(2) uses decomposed matrices $DM_{m \times r}$ and $DM_{r \times n}$ as affine transform component in network. The input is one-hot vector and output is the corresponding column of the original matrix.

\begin{equation}
	OUT= DM_{m \times r} DM_{r \times n} IN_{oneHot}
	\label{eq1}
\end{equation}

The index of 1 in $IN_{oneHot}$ is corresponding to column index of destination matrix. 
Network (1) requires large computation resources because of its large size of input and output.

\begin{equation}
	INPARNUM = NR \times NL
	\label{eq1}
\end{equation}

\begin{equation}
	OUTPARNUM = RANK \times (NR + NL)
	\label{eq1}
\end{equation}

where INPARNUM, OUTPARNUM, NR, NL and RANK is input dimension, output dimension, original matrix rows, original matrix column and destination rank. It needs a vast resource to compute.  So we decide to implement network (2).

Actually, this operation is similar to bottleneck and sparse operation is equivalent to the pre-training process. 

\section{EXPERIMENT}
\label{sec:typestyle}

We implement our sparse algorithm in KWS problem. We use Google Speech Command Dataset [16] as train and test data. We choose happy (1742 utterances) as keyword and use a 23h environment data to test FA. To get a robust model, we mix noise into original voice data. And we use KALDI toolkit [17] to train each model in our experiment. 

We use 3-layer and 4-layer DNN as baseline models. Each model has 858 input dimension and 3 output dimension. 

%\subsection{Tables}

%n example of a table is shown in Table~\ref{tab:example}. The caption text must be above the table.

%\begin{table}[th]
 % \caption{This is an example of a table}
%  \label{tab:example}
 % \centering
%  \begin{tabular}{ r@{}l  r }
   % \toprule
 %   \multicolumn{2}{c}{\textbf{}} & 
%                                         \multicolumn{1}{c}{\textbf{Decibels}} \\
%    \midrule
 %   $1$                       & $/10$ & $-20$~~~             \\
   % $1$                       & $/1$  & $0$~~~               \\
  %  $2$                       & $/1$  & $\approx 6$~~~       \\
  %  $3.16$                    & $/1$  & $10$~~~              \\
  %  $10$                      & $/1$  & $20$~~~              \\
 %   $100$                     & $/1$  & $40$~~~              \\
%    $1000$                    & $/1$  & $60$~~~              \\
%    \bottomrule
%  \end{tabular}
  
%\end{table}

\subsection{Sparse Algorithm}
\label{ssec:subhead}
	We apply Sparse Algorithm 4.1.1, 4.1.2 and 4.1.3 for each baseline model of each keyword respectively.

Then we use environment and test dataset to evaluate the True Alarm (TA) and False Alarm (FA) of each model. And the ROC is showed in Figure.6.

\begin{figure}[htb]

%\begin{minipage}[b]{1.0\linewidth}
% \centering
%  \centerline{\includegraphics[width=8.5cm]{test.jpeg}}
 % \vspace{2.0cm}
 % \centerline{(a) Result 1}\medskip
%\end{minipage}

\begin{minipage}[b]{.48\linewidth}
  \centering
  \centerline{\includegraphics[width=4.5cm]{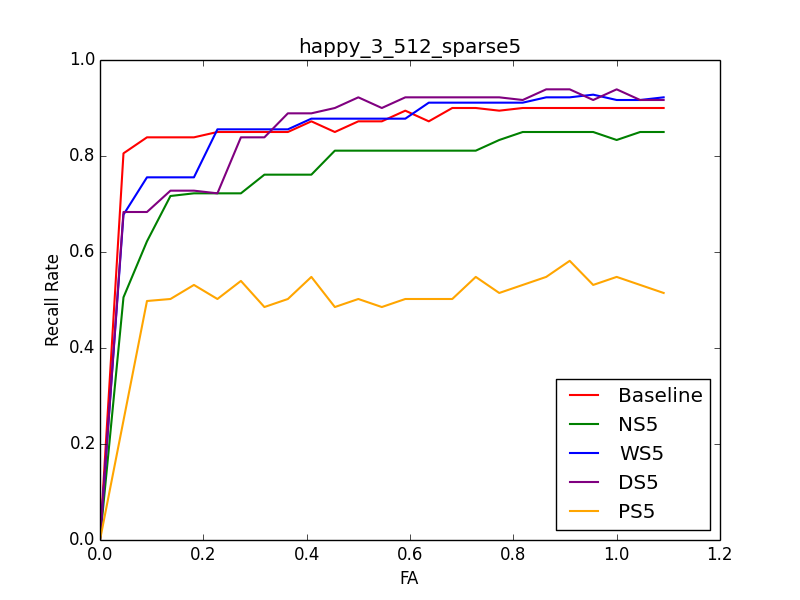}}
 % \vspace{1.5cm}
 \centerline{(1)  }\medskip
\end{minipage}
%!TEX encoding = UTF-8 Unicode%\hfill
\begin{minipage}[b]{0.48\linewidth}
 \centering
  \centerline{\includegraphics[width=4.5cm]{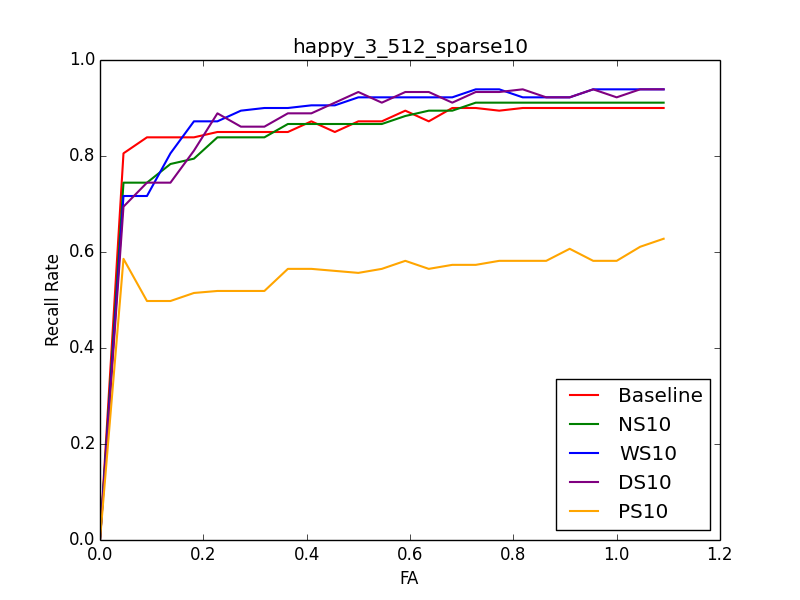}}
 % \vspace{1.5cm}
  \centerline{(2)  }\medskip
\end{minipage}

\begin{minipage}[b]{.48\linewidth}
  \centering
  \centerline{\includegraphics[width=4.5cm]{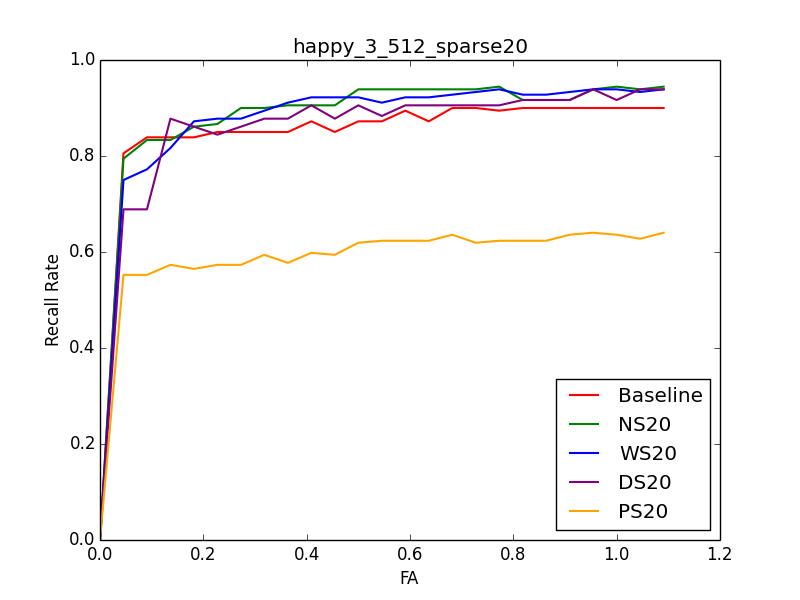}}
 % \vspace{1.5cm}
 \centerline{(3)  }\medskip
\end{minipage}
%!TEX encoding = UTF-8 Unicode%\hfill
\begin{minipage}[b]{0.48\linewidth}
 \centering
  \centerline{\includegraphics[width=4.5cm]{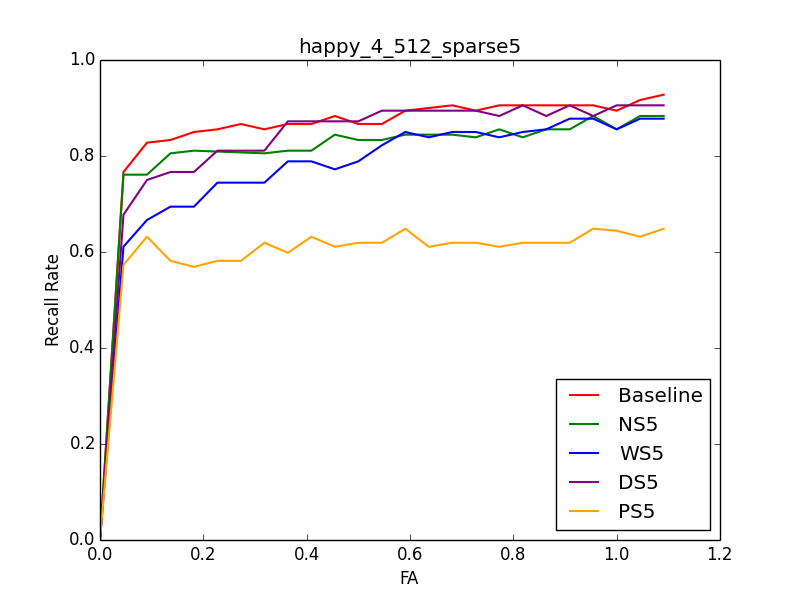}}
 % \vspace{1.5cm}
  \centerline{(4)  }\medskip
\end{minipage}

\begin{minipage}[b]{.48\linewidth}
  \centering
  \centerline{\includegraphics[width=4.5cm]{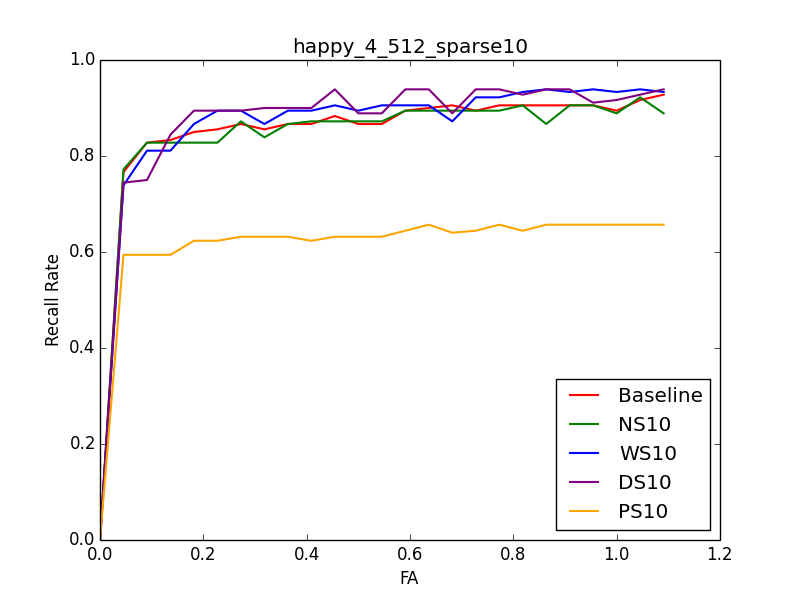}}
 % \vspace{1.5cm}
 \centerline{(5)  }\medskip
\end{minipage}
%!TEX encoding = UTF-8 Unicode%\hfill
\begin{minipage}[b]{0.48\linewidth}
 \centering
  \centerline{\includegraphics[width=4.5cm]{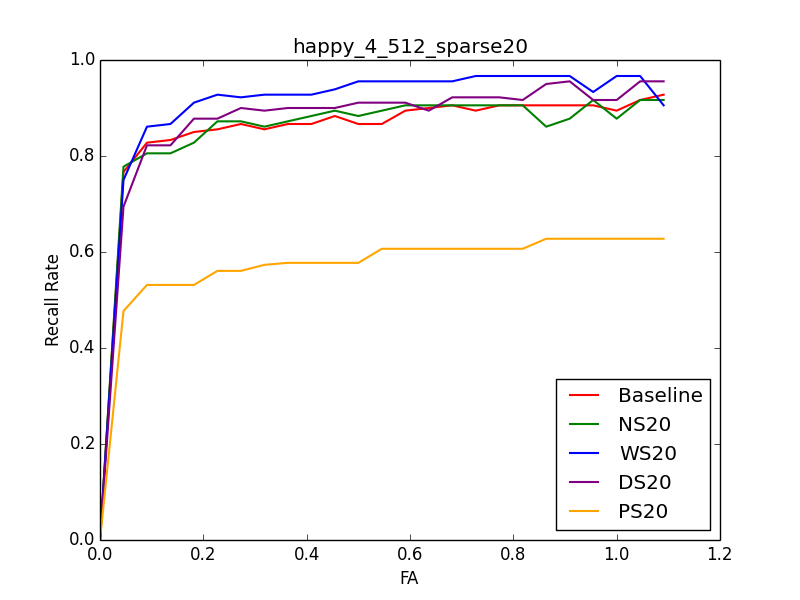}}
 % \vspace{1.5cm}
  \centerline{(6)  }\medskip
\end{minipage}

\caption{Sparse model performance. (1) (2) (3) are based on $3 \times 512$ model with 5\% 10\% 20\% remain parameters. (4) (5) (6) are  based on $4 \times 512$ model with 5\% 10\% 20\% remain parameters. Baseline is a fully connected network. NS is based on 4.1.1, WS is based on 4.1.2, DS is based on 4.1.3, PS is full connect network whose total parameters equals to sparse model}%
\label{fig:res}

\end{figure}

From (1) - (6), the performance of sparse model does not decline much. In some situation, the sparse model even performs better, especially with large parameter remain rate. And all sparse model performs much better than full connected model whose total parameters equals to sparse model.  The reason may be that the topology structure difference between sparse model and fully connected network. Although the sparse model has a rare number of parameter, its node number is still large. This property keeps high information reservation and emphasizes important part. 

\subsection{Matrix Decomposition}
\label{ssec:subhead}
We have decomposed the matrices in affine component of ``happy" A1 model. The performance declines obviously. Its FA is almost 10 when recall rate is 0.93. 

We assume two main reasons cause the result. As we mention before, We use matrices $DM_{m \times r}$ and $DM_{r \times n}$ to replace sparse matrix $M_{m \times n}$.  We calculate the rank of $DM_{r \times n}$ and most of them are full rank. But the rank of  $DM_{m \times r}$ and $DM_{r \times n}$ production is not greater than $r$. So this operation leads loss into the model. Lower rank, higher loss. Another reason is insufficient training data. The model has tens of thousands parameter. The training data may be insufficient to train. We test the frame accuracy of each model and it does not decrease much. So the overfitting is serious. The frame accuracy of each model is shown in Table.1.

\begin{table}[th]

  \centering
  \begin{tabular}{ cl  c }
 %   \toprule
    \multicolumn{1}{c}{\textbf{Model}} & 
                                         \multicolumn{1}{c}{\textbf{Frame Acc.}} \\
%    \midrule
    $3 \times 512$                  & $\quad 94.42\%$~~~             \\
    $3 \times 512 (DCS5)$                  & $\quad 93.35\%$~~~             \\
    $3 \times 512 (DCS10)$                  & $\quad 93.95\%$~~~             \\
    $3 \times 512 (DCS20)$                  & $\quad 94.18\%$~~~             \\
    $4 \times 512$                  & $\quad 94.35\%$~~~             \\
    $4 \times 512(DCS5)$                  & $\quad 93.48\%$~~~             \\
    $4 \times 512(DCS10)$                  & $\quad 93.96\%$~~~             \\
    $4 \times 512(DCS20)$                  & $\quad 94.17\%$~~~             \\
  %  \bottomrule
  \end{tabular}
    \caption{Frame accuracy of each model}
  \label{tab:example}
\end{table}

Then we do this experiment on our keyword. We use more than 50K utterances to train the model and test the result. 

It performs better than ``happy" obviously. Moreover, the decomposition sparse models even do not re-converge because of its long computation time. The performance is shown in Figure.7. Models' size is shown in Table.2

\begin{figure}[htb]

\begin{minipage}[b]{1.0\linewidth}
 \centering
  \centerline{\includegraphics[width=8.5cm]{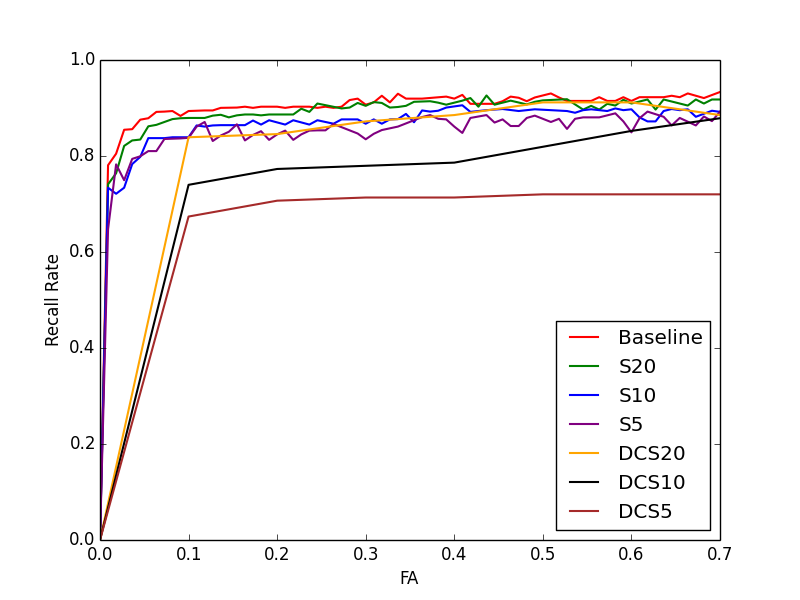}}
 % \vspace{2.0cm}
%  \centerline{(a) Result 1}\medskip
\end{minipage}

\caption{Sparse model performance. DCS is decomposition sparse model}%
\label{fig:res}

\end{figure}

%\begin{figure}[htb]
%
%\begin{minipage}[b]{1.0\linewidth}
% \centering
%  \centerline{\includegraphics[width=8.5cm]{hinomiCompareSparsewithDecomp2.png}}
% % \vspace{2.0cm}
%%  \centerline{(a) Result 1}\medskip
%\end{minipage}
%
%\caption{FST for keyword ``happy"}%
%\label{fig:res}
%
%\end{figure}

%\begin{figure}[htb]
%
%\begin{minipage}[b]{1.0\linewidth}
% \centering
%  \centerline{\includegraphics[width=8.5cm]{hinomiCompareSparsewithPara3.png}}
% % \vspace{2.0cm}
%%  \centerline{(a) Result 1}\medskip
%\end{minipage}
%
%\caption{FST for keyword ``happy"}%
%\label{fig:res}
%
%\end{figure}

\begin{table}[th]

  \centering
  \begin{tabular}{ cl  c }
 %   \toprule
    \multicolumn{1}{c}{\textbf{Model}} & 
                                         \multicolumn{1}{c}{\textbf{Model Size}} \\
%    \midrule
    $4 \times 512$                  & $\quad 4.7M$~~~             \\
    $4 \times 512 (DCS5)$                  & $\quad 326K$~~~             \\
    $4 \times 512 (DCS10)$                  & $\quad 491K$~~~             \\
    $4 \times 512 (DCS20)$                  & $\quad 967K$~~~             \\
  \end{tabular}
    \caption{Model size of each model}
  \label{tab:example}
\end{table}

When remain rate is high enough, DCS performs relative good. The model size is much smaller than original baseline model. It is worth to be considered when the model has to be implemented on some source-short device such as car tablet. 

But when remain rate is lower, DCS performs bad.  Besides not re-converge, the bottleneck topology causes a significant loss to the network.

\section{CONCLUSION}
\label{sec:majhead}

We have implemented sparse model into KWS problem, and the performance is acceptable. And the result of the sparse model can be used to guide model selection. But how to implement it into the real project is next challenge. Matrix Decomposition may lead significant loss into the model. So its application scenarios are restricted. We will try to find a method to realize sparse model in general ASR system in the future. Besides this, we will also do more research on implementing the sparse algorithm on other models such as CNN and LSTM.

\section{REFERENCE}
\label{sec:majhead}

\noindent[1] Xin Li and Changsong Liu, ``Prune the Convolutional Neural Networks With Sparse Shrink",  arXiv: 1708.02439.

\noindent[2] Song Han, Jeff Pool, John Tran and Wiliiam J.Dally, ``Learning Both Weights And Connections For Efficient Neural Networks", \emph{Advances in Neural Information Processing System (NIPS)}, December 2015.

\noindent[3] Song Han, Huizi mao and William J.Dally, ``Deep Compression: Compressing Deep Neural Network with Pruning, Trained Quantization and Huffman Coding", \emph{International Conference on Learning Representations(ICLR)}, May 2016.

\noindent[4] Sharan Narang, Erich Elsen, Greg Diamos and Shubho Sengupta. ``Exploring Sparsity in Recurrent Neural Networks", arXiv: 1704.05119.

\noindent[5] Sharan Narang, Eric Undersander and Gregory Diamos, ``Block-Sparse Recurrent Neural Networks", arXiv: 1711.02782.

\noindent[6] Babak Hassibi and David G Stork. ``Second Order Derivatives for Network Pruning: Optimal Brain Surgeon", \emph{Morgan Kaufmann} 1993.

\noindent[7] Yann Le Cun, John S, Denker and Sara A. Solla, ``Optimal Brain Damange".

\noindent[8] David RH Miller, Michael Kleber, Chia-Lin Kao, Owen Kimball, Thomas Colthurst, Stephen A Lowe, Richard M Schwartz and Herbert Gish, ``Rapid and accurate spolen term detection", in \emph{Eighth Annual Conference of the International Speech Communication Association} 2007

\noindent[9] Siddika Parlak and Murat Saraclar, ``Spoken term detection for turkish broadcast news", in \emph{Acoustics, Speech and Signal Processing},  2008, pp. 5244-5247

\noindent[10] Richard C Rose and Douglas B Paul, `` A Hidden Markov Model Based Keyword Recognition System", in \emph{Acoustics, Speech, and Signal Processing, 1990. ICASSP-90., 1990 International Conference on.} IEEE, 1990, pp. 129-132.

\noindent[11] Jay G Wilpon, Lawrence R Rabiner, C-H Lee, and ER Goldman, ''Automatic Recognition of Keywords in Unconstrained Speech Using Hidden Markov Model", \emph{IEEE Transactions on Acoustics, Speech and Signal Processing}, vol. 38, no.11, pp. 1870-1878, 1990.

\noindent[11] Geoffrey Hinton, Li Deng, Dong Yu, George E Dahl, Abdel-rahman Mohamed, Navdeep Jaitly, Andrew Senior, Vincent Vanhoucke, Patrick Nguyen, Tara N Sainath, et al., `` Deep Neural Networks for Acoustic Modeling in Speech Recognition: The Shared Views of Four Research Groups", \emph{IEEE Signal Processing Magazine}, vol. 29, no.6, pp.82-97, 2012.

\noindent[13] Sankaran Panchapagesan, Ming Sun, Aparna Khare, Spyros Matsoukas, Arindam Mandal, Bj\"orn Hoffmeister and Shiv Vitaladevuni, ``Multi-task Learning and Weighted Cross-entropy for DNN-based Keyword Spotting", \emph{Interspeech 2016}, pp.760-764, 2016.

\noindent[14] Ming Sun, David Snyder, Yixin Gao, Varun Nagaraja, Mike Rodehorst, Sankaran Panchapagesan, Nikko Strom, Spyros Matsoukas and Shiv Vitaladevuni, ``Compressed Time Delay Neural Network for Small Footprint Keyword Spotting", \emph{Proc. Interspeech 2017}, pp. 3607-3611, 2017.

\noindent[15] Guoguo Chen, Sanjeev Khudanpur, Daniel Povey, Jan Trmal, David Yarowsky and Oguz Yilmaz, ``Quantifying the Value of Pronunciation Lexicons for Keyword Search In Low Resource Languages", \emph{IEEE international Conference on Acoustics}, 2013.

\noindent[16] Pete Warden, ``Speech Commands: A Data Set for Limited-Vocabulary Speech Recognition", arXiv: 1804.03209

\noindent[17] Daniel Provey et al. ``The Kaldi Speech Recognition Toolkit", in \emph{IEEE AERU}. IEEE Signal Processing Society, 2011, number EPFL-CONF-192584
\bibliographystyle{IEEEbib}
\bibliography{strings,refs}

\end{document}